\RequirePackage{amsthm}  
\documentclass[sn-vancouver,Numbered]{sn-jnl}

\usepackage{graphicx}%
\usepackage{multirow}%
\usepackage{amsmath,amssymb,amsfonts}%
\usepackage{amsthm}%
\usepackage{mathrsfs}%
\usepackage[title]{appendix}%
\usepackage{xcolor}%
\usepackage{textcomp}%
\usepackage{manyfoot}%
\usepackage{booktabs}%
\usepackage{algorithm}%
\usepackage{algorithmicx}%
\usepackage{algpseudocode}%
\usepackage{listings}%
\usepackage{hyperref}
\usepackage{url}
\usepackage{siunitx}
\usepackage{subcaption}
\usepackage{layouts}
\usepackage{tikz}
\usepackage{lmodern}
\usepackage{anyfontsize}
\usetikzlibrary{positioning}
\usepackage{acronym}
\newcommand{\lpbf}{\mbox{PBF-LB/M} } 

\theoremstyle{thmstyleone}%
\theoremstyle{thmstyletwo}%

\theoremstyle{thmstylethree}%

\begin{document}

\def\authorHAZ{Hans Aoyang Zhou}
\def\authorMK{Marco Kemmerling}
\def\authorAA{Anas Abdelrazeq}
\def\authorRS{Robert H. Schmitt}

\def\authorJT{Jan Theunissen}
\def\authorJHS{Johannes Henrich Schleifenbaum}

\title[Estimating Pore Location of \lpbf Processes with Segmentation Models]{Estimating Pore Location of \lpbf Processes with Segmentation Models}


\author*[1]{\fnm{Hans Aoyang} \sur{Zhou}}\email{hans.zhou@wzl-iqs.rwth-aachen.de}

\author[2]{\fnm{Jan} \sur{Theunissen}}\email{jan.theunissen@dap.rwth-aachen.de}
\equalcont{These authors contributed equally to this work.}

\author[1]{\fnm{Marco} \sur{Kemmerling}}\email{marco.kemmerling@wzl-iqs.rwth-aachen.de}
\equalcont{These authors contributed equally to this work.}

\author[1]{\fnm{Anas} \sur{Abdelrazeq}}\email{anas.abdelrazeq@wzl-iqs.rwth-aachen.de}

\author[2]{\fnm{Johannes Henrich} \sur{Schleifenbaum}}\email{johannes.henrich.schleifenbaum@dap.rwth-aachen.de}

\author[1]{\fnm{Robert H.} \sur{Schmitt}}\email{robert.schmitt@wzl-iqs.rwth-aachen.de}

\affil*[1]{\orgdiv{Laboratory for Machine Tools and Production Engineering}, \orgname{RWTH Aachen University}, \orgaddress{\street{Campus Boulevard 30}, \city{ Aachen}, \postcode{52074}, \country{Germany}}}

\affil[2]{\orgdiv{Digital Additive Production DAP}, \orgname{RWTH Aachen University}, \orgaddress{\street{Campus Boulevard 73}, \city{Aachen}, \postcode{52074}, \country{Germany}}}


\abstract{
Reliably manufacturing defect free products is still an open challenge for Laser Powder Bed Fusion processes. 
Particularly, pores that occur frequently have a negative impact on mechanical properties like fatigue performance.
Therefore, an accurate localisation of pores is mandatory for quality assurance, but requires time-consuming post-processing steps like computer tomography scans. 
Although existing solutions using in-situ monitoring data can detect pore occurrence within a layer, they are limited in their localisation precision. 
Therefore, we propose a pore localisation approach that estimates their position within a single layer using a Gaussian kernel density estimation.
This allows segmentation models to learn the correlation between in-situ monitoring data and the derived probability distribution of pore occurrence.
Within our experiments, we compare the prediction performance of different segmentation models depending on machine parameter configuration and geometry features. 
From our results, we conclude that our approach allows a precise localisation of pores that requires minimal data preprocessing. 
Our research extends the literature by providing a foundation for more precise pore detection systems. 
}




\maketitle


\section{Introduction}
\label{sec:introduction}


Although Laser Powder Bed Fusion (PBF-LB/M) has proven to perform well in a wide range of applications, its missing reproducibility limits the expansion to new application areas. As long as quality requirements can not be met consistently, \lpbf will remain a manufacturing process for niche applications like biomedical \cite{Arif.2023} or aerospace \cite{BlakeyMilner.2021} domains. Because stochastically occurring defects are the main cause preventing reproducibility, a reliable defect detection system would ensure high product quality. Therefore, researchers have studied the benefit of in-situ monitoring to leverage the layerwise manufacturing principle to detect defects as early as possible \cite{Grasso.2021}. Advancements in the Industrial Internet of Things like Artificial Intelligence \cite{Behery.2023} and Digital Shadows \cite{Liebenberg.2023}, that combine implicit engineering knowledge with data-driven approaches to generate actionable knowledge for either machines or decision-makers, push the development of a closed-loop control system that potentially achieves defect-free fabrication outcomes closer to reality. 

One of the most common defect types in additively manufactured metal parts are pores, which usually results from a suboptimal configuration of laser power. They have a significant impact on mechanical properties like fatigue strength and fatigue life \cite{Sanaei.2021}. Consequently, due to their common occurrence and critical influence on functional properties, their formation, and detection, was studied frequently in the literature. Therefore, previous attempts tried to predict porosity using a wide range of monitoring data types. An overview of related references are listed in \autoref{tab:related_work}, where we compare each contribution regarding data source, labelling approach, model type and localization precision. From the comparison, we deduce the following general limitations in detecting pores for \lpbf processes.

\begin{table}[p]
\centering
\caption{Related Work with their data sources, labelling method, model type and prediction task used.}
\label{tab:related_work}
\begin{tabular}{c|ccccc|ccc|ccc|ccc|}
\toprule
 & \multicolumn{5}{c|}{Data Source} & \multicolumn{3}{c|}{Labeling} & \multicolumn{3}{c|}{Model} & \multicolumn{3}{c|}{Task} \\ \cmidrule(lr){2-6} \cmidrule(lr){7-9} \cmidrule(lr){10-12} \cmidrule(r){13-15}
Ref. &
  \rotatebox{90}{Visible Light HRC} &
  \rotatebox{90}{Pyrometer} &
  \rotatebox{90}{Acoustic} &
  \rotatebox{90}{X-Ray CT} &
  \rotatebox{90}{Photodiodes} &
  \rotatebox{90}{X-Ray CT} &
  \rotatebox{90}{Energy Density} &
  \rotatebox{90}{Geometrical Flaws} &
  \rotatebox{90}{Classical ML} &
  \rotatebox{90}{MLP} &
  \rotatebox{90}{CNN} &
  \rotatebox{90}{Binary Classification} &
  \rotatebox{90}{Multiclass Classification} &
  \rotatebox{90}{Segmentation} \\ \midrule
\cite{Shevchik.2018}    &            &            & \checkmark &            &            &            & \checkmark &            &            &            & \checkmark &            & \checkmark &            \\
\cite{Khanzadeh.2018}   &            & \checkmark &            &            &            & \checkmark &            &            & \checkmark &            &            & \checkmark &            &            \\
\cite{Gobert.2018}      & \checkmark &            &            &            &            & \checkmark &            &            & \checkmark &            &            & \checkmark &            &            \\
\cite{Zhang.2019}       & \checkmark &            &            &            &            & \checkmark &            &            &            &            & \checkmark & \checkmark &            &            \\
\cite{Seifi.2019}       &            & \checkmark &            &            &            & \checkmark &            &            & \checkmark &            &            & \checkmark &            &            \\
\cite{Imani.2019}       & \checkmark &            &            &            &            & \checkmark &            & \checkmark &            &            & \checkmark & \checkmark &            &            \\
\cite{Gaikwad.2019}     & \checkmark &            &            &            &            & \checkmark &            &            &            &            & \checkmark &            & \checkmark &            \\
\cite{Caggiano.2019}    & \checkmark &            &            &            &            &            & \checkmark &            &            &            & \checkmark &            & \checkmark &            \\
\cite{Kwon.2020}        & \checkmark &            &            &            &            &            & \checkmark &            &            & \checkmark &            &            & \checkmark &            \\
\cite{Gobert.2020}      &            &            &            & \checkmark &            & \checkmark &            &            &            &            & \checkmark &            &            & \checkmark \\
\cite{Snow.2021}        & \checkmark &            &            &            &            & \checkmark &            &            &            &            & \checkmark & \checkmark &            &            \\
\cite{Petrich.2021}     & \checkmark &            & \checkmark &            & \checkmark & \checkmark &            &            &            & \checkmark &            & \checkmark &            &            \\
\cite{Zhang.2022}       &            &            &            & \checkmark &            & \checkmark &            &            &            &            & \checkmark &            &            & \checkmark \\
\cite{Yang.2022}        &            &            &            & \checkmark &            & \checkmark &            &            &            &            & \checkmark &            &            & \checkmark \\
\cite{Snow.2022}        & \checkmark &            & \checkmark &            & \checkmark & \checkmark &            &            &            & \checkmark & \checkmark & \checkmark &            &            \\
\cite{Pandiyan.2022}    &            &            & \checkmark &            & \checkmark & \checkmark &            &            &            &            & \checkmark &            & \checkmark &            \\
\cite{Nemati.2022}      &            &            &            & \checkmark &            &            &            &            &            &            & \checkmark &            &            & \checkmark \\
\cite{Ansari.2022}      & \checkmark &            &            &            &            & \checkmark &            & \checkmark &            &            &            & \checkmark &            &            \\
\cite{Ye.2023}          &            &            &            & \checkmark &            & \checkmark &            &            & \checkmark &            &            &            & \checkmark &            \\
\cite{Surana.2023}      &            &            &            &            & \checkmark & \checkmark & \checkmark &            &            &            & \checkmark &            &            & \checkmark \\
\cite{Gorgannejad.2023} &            &            & \checkmark &            & \checkmark & \checkmark &            &            & \checkmark &            &            & \checkmark &            &            \\
\cite{Pan.2024}         &            &            &            & \checkmark &            & \checkmark &            &            &            &            & \checkmark &            &            & \checkmark \\
\cite{Dong.2024}        &            &            &            & \checkmark &            & \checkmark &            &            &            &            & \checkmark &            &            & \checkmark \\
\cite{Desrosiers.2024}  &            &            &            & \checkmark &            & \checkmark &            &            &            &            & \checkmark &            &            & \checkmark \\ \midrule
Ours                    & \checkmark &            &            &            &            & \checkmark &            &            &            &            & \checkmark &            &            & \checkmark \\
\bottomrule
\end{tabular}
\end{table}

First, the majority of contributions simplify pore locations to a single quantity (e.g., porosity value). These simplifications cover either discrete porosity levels \cite{Shevchik.2018}, binary pore occurrence within a predefined area — most commonly within a layer — \cite{Khanzadeh.2018, Gobert.2018, Zhang.2019, Seifi.2019, Imani.2019, Snow.2021, Petrich.2021, Snow.2022, Pandiyan.2022, Ansari.2022, Ye.2023, Gorgannejad.2023} or indirectly through other quality parameters \cite{Gaikwad.2019, Caggiano.2019, Kwon.2020}. In other words, these approaches omit the required precision to localize pores within a single layer.     
To the best of our knowledge, the only publications that localizes pores with Machine Learning within a single layer, uses cross-section Computer Tomography (CT) images as source data \cite{Gobert.2020, Zhang.2022, Yang.2022, Nemati.2022, Pan.2024, Dong.2024, Desrosiers.2024}. However, CT cross-section images are captured off-line, and their use result in limited part size and manufacturing speed. Only one work localizes pores using in-situ monitoring data from two photodiodes \cite{Surana.2023}. Currently, the precision of pore localization within in-situ monitoring data is limited to the resolution of a single layer. In contrast, we aim to improve pore localization precision, by predicting pore occurrence on pixel-level within a single layer of in-situ monitoring data.

Second, we further deduce from \autoref{tab:related_work} that a variety of monitoring sensors with different data modalities were investigated to predict porosity. The most common ones are off-axial high-resolution cameras within the spectrum of visible light, photodiodes, pyrometers, acoustic sensors, or a combination of those. Localizing pores using in-situ visual monitoring data alone was not attempted, instead only off-line visual data (e.g. CT data) was used for segmenting pores. In our work, we combine the data from two off-axial camera sources, one within the spectrum of visible and another one near-infrared light camera source. Combining different sensor types usually requires extensive transformations and manual feature extraction steps to unify the data modality (e.g. time series to image \cite{Surana.2023}). A sophisticated data transformation and feature extraction pipeline hamper its transfer to industrial applications due to higher required computing resources and increased reproducibility difficulty.

Third, purposefully induced defects are necessary to ensure model performance. Because the appearance of pores is not always deterministic, some work induces defects on purpose through either machine parameter configuration \cite{Shevchik.2018, Caggiano.2019, Kwon.2020, Surana.2023} or geometry design \cite{Imani.2019, Ansari.2022}. Inducing defects on purpose introduces a bias within the dataset, therefore simplifying the pore localization task significantly. These configurations would not be present in realistic production circumstances, therefore preventing a direct transfer to industrial standards.

We structure our contribution into the following sections:
\begin{enumerate}
    \item We introduce a new simplification approach by reformulating the pore localization to a pore estimation problem (cf. sections \ref{subsec:data_labeling} and \ref{subsec:ppp_with_seg}).
    \item We apply different segmentation models to estimate pore locations and evaluate their performance dependent on geometry features and process parameters (cf. section \ref{sec:results}).
    \item We reflect on the implications of our results and argue that it is a simple but effective approach to precisely localize pores (cf. section \ref{sec:discussion}) 
\end{enumerate}

\section{Pore Location Estimation}
\label{sec:methods}
The limitations of the presented related work, indicate that current deep learning approaches are not able to accurately localize pores in an image of a \lpbf layer captured by a camera that operates in visible to near-infrared range. Possible reasons may be the stochastic nature of pore occurrence or the limited modelling capacity of current state-of-the-art deep learning models. Therefore, we propose to relax the localization task of precisely allocating pores at pixel location, by an estimation task of predicting the probability that pores occur at pixel location. 

In fact, we derive for each layer, based on determined pore locations, a probability distribution that estimates the occurrence of pores; Given the pore positions $\left\{\mathbf{v}_i \mid \mathbf{v}_i \in \mathbb{R}^2, 1 \leq i \leq Q \right\}$ with $Q$ as the total number of pores in an image, we derive a probability distribution of pore occurrence for each pixel position $\mathbf{v}$ using a kernel density estimation (KDE)  
\begin{equation}
\label{eq:kde}
    \mathit{KDE}(\mathbf{v}) = \frac{1}{\beta Q} \sum_{i=1}^{Q} N\left(\frac{\mathbf{v}-\mathbf{v_i}}{\beta}\right),
\end{equation}
where $N \left( \cdot \right)$ is selected to be a Gaussian probability density function and bandwidth $\beta = 20$. We selected $\beta$ by inspecting the resulting pore probabilities visually so that it matches a plausible probability distribution. We scale the probability distribution to $[0, 1]$ using min-max normalization and store it as a grayscale image $x_{PP}$. By replacing the discrete segmentation masks, the derived pore location estimations are used as labels for training.

We validate our approach of pore location estimation regarding applicability by testing and comparing a variety of segmentation models to learn the correlation between in-situ monitoring data and pore occurrence probability. For that, we fabricate different specimen with changing process parameter configuration and geometries to test the robustness of our approach under different fabrication conditions. 


\section{Experimental Design}
\label{sec:experiments}
With our pore estimation approach defined, we aim to validate the applicability and robustness of our approach. For that, we first designed and fabricated two different test geometries covering different geometry features and machine parameter configurations (cf. section \ref{subsec:part_design}). Afterward, we collect layerwise in-situ monitoring data during processing and derive pore locations during post-processing (cf. section \ref{subsec:monitoring_setup}). Within the fabricated specimen, we use an X-ray computer tomograph to scan them for pores (cf. section \ref{subsec:XCT_measurements}). Next, the acquired data is integrated layerwise into a coherent dataset that matches monitoring data with pore probability distribution (cf. section \ref{subsec:data_labeling}). With the derived dataset, we train different segmentation models to predict the pore probability distribution given the captured monitoring data (cf. section \ref{subsec:ppp_with_seg}).

\subsection{Part design}
\label{subsec:part_design}
\begin{figure}[htb]
    \begin{subfigure}{0.33\textwidth}
        \centering
        \includegraphics[width=\textwidth]{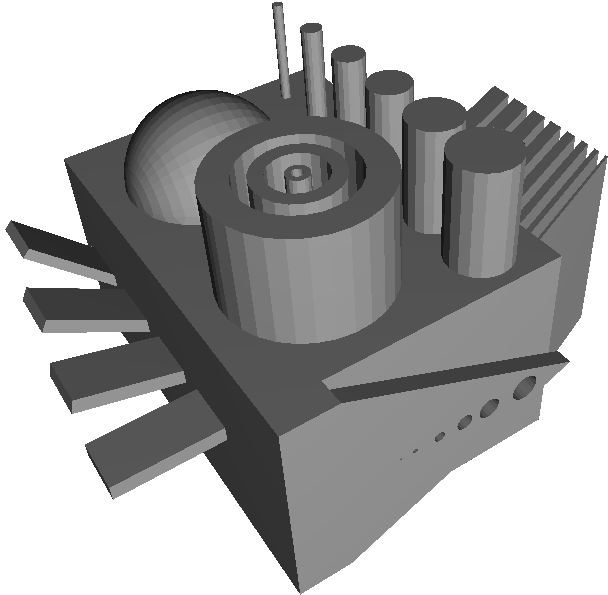}
        \caption{}
        \label{fig:KS_geometry}
    \end{subfigure}
    \begin{subfigure}{0.3\textwidth}
        \centering
        \includegraphics[width=\textwidth]{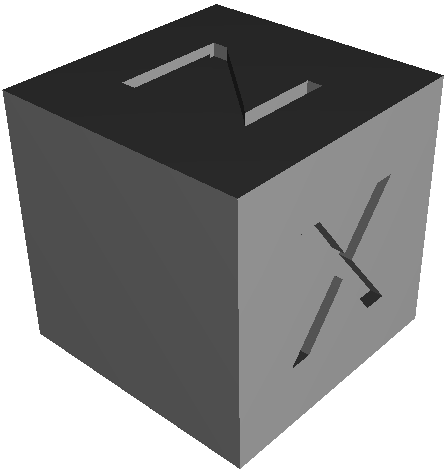}
        \caption{}
        \label{fig:XYZ_cube}
    \end{subfigure}  
    \hfill
    \begin{subfigure}{0.33\textwidth}    
        \centering
        \includegraphics[width=0.9\textwidth]{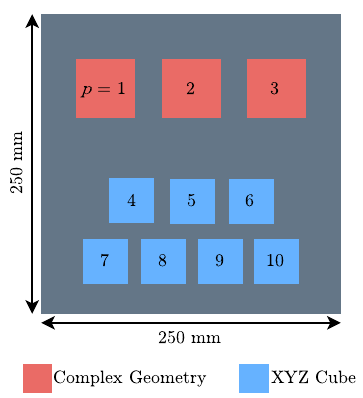}
        \caption{}
        \label{fig:part_positions}
    \end{subfigure}
    \caption{Geometries selected for our experiments are (a) Complex Geometry and (b) XYZ Cube. (c) shows the relative arrangement of part positions $p = 1, \dots , \Pi$ within our building platform.}
    \label{fig:geomtries}
\end{figure}

Two parts with different geometric complexity characteristics have been designed to validate our modelling approach. They are shown in \autoref{fig:geomtries}. The selection of the complex geometry shown in \autoref{fig:KS_geometry} serves two main purposes. Firstly, the complex geometry of the part distributes the probability of defects unevenly over different geometry shapes. Secondly, the model is tested on the ability to generalize through the use of different geometric cross-sections. The components with simpler geometric features serve to generalize the model by mapping different machine parameter configurations with different process parameters. At the same time, the effect of cross-sectional changes between the layers is minimized.
An XYZ cube shown in \autoref{fig:XYZ_cube} was chosen as a representative for the simple geometry. This design is a common test geometry that has a simple geometry structure and allows the alignment of spatial orientation between different data sources (i.e. HR and CT). The dimensions of the cube with a size of $20 \times 20 \times 20$ \si{\milli\meter} are constrained by the dimensions of the hardware used for the computer tomography images. Both geometries are manufactured in the same printing job with the arrangement on the building platform displayed in \autoref{fig:part_positions}.

\subsection{In-situ monitoring setup}
\label{subsec:monitoring_setup}
\begin{figure}
     \centering
     \begin{subfigure}[b]{0.45\textwidth}
         \centering
         \includegraphics[width=\textwidth]{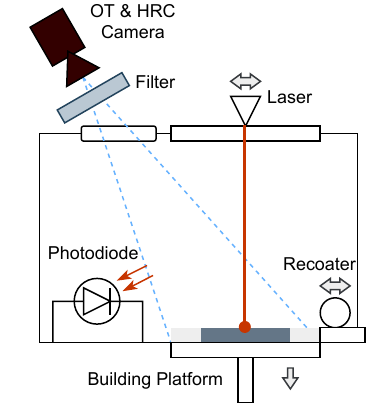}
         \caption{Schematic setup}
         \label{fig:OT_monitoring_setup_schema}
     \end{subfigure}
     \hfill
     \begin{subfigure}[b]{0.45\textwidth}
         \centering
         \includegraphics[width=\textwidth]{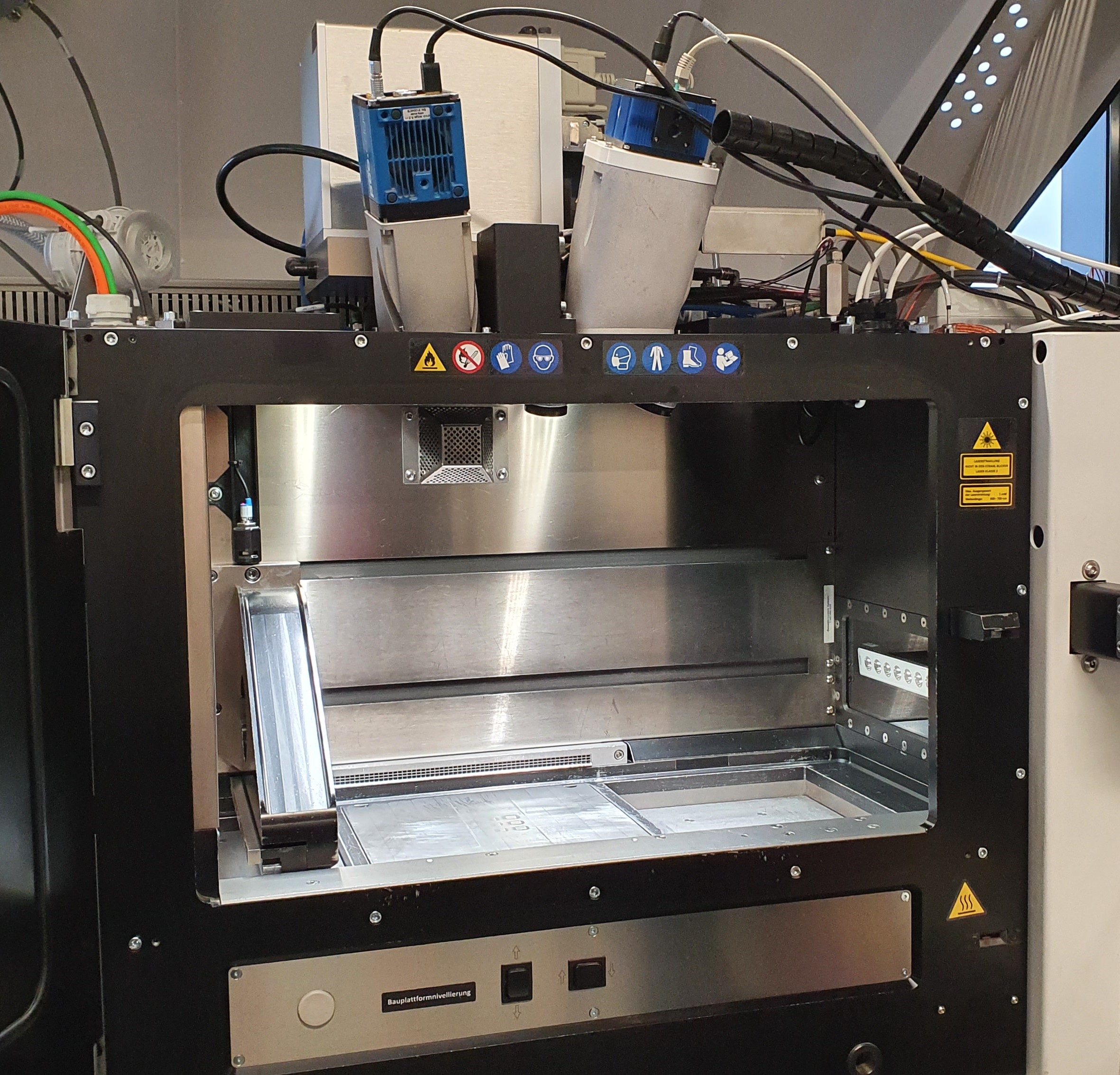}
         \caption{Real-world setup}
         \label{fig:OT_monitoring_setup_DAP}
     \end{subfigure}
     \caption{\lpbf experiment and monitoring setup, displayed (a) schematically with its corresponding (b) real-world implementation at the Digital Additive Production chair (DAP).}
     \label{fig:OT_monitoring_setup}
\end{figure}

The used machine setup 
uses an \textit{EOS M 290} machine (\textit{EOS GmbH, Germany}) system with two integrated sensors. The monitoring setup is depicted in \autoref{fig:OT_monitoring_setup}.
First, an Optical Tomography (OT) system \footnote{The presented OT monitoring setup is provided by EOS since 2020 as an optional commercially available software solution named EOSTATE Exposure OT \cite{Fuchs.2020}} proposed by \cite{Zenzinger.2015} that uses a \textit{pco edge 5.5 \mbox{sCMOS}} camera combined with a band-pass filter within near-infrared range to filter only thermally relevant emissions. 
The OT system captures layer-wise thermal radiation signatures with a resolution of  $2560 \times \SI{1060}{px}$. In-situ monitoring systems using near-infrared images have shown evidence to correlate with defect location like pores \cite{Yoder.2019} or lack of fusion \cite{Gogelein.2018}.
Second, a high-resolution (HR) camera \textit{SVS-Vistek hr29050MFLGEC} is integrated into the building chamber. It operates in the wavelength of visible light, with a total resolution of $6576 \times \SI{4384}{px}$. Although the OT system continuously captures multiple images per layer, we sum up the intensities of all images in one layer into one image per layer. The HR images are captured after a layer is fabricated and before the recoating of a new layer. With an integrated photodiode into the machine setup, we detect the fabrication start and end of each layer. This allows a clear matching of all images with their corresponding layer number. 

\subsection{Sample fabrication}
\label{subsec:Sample_fabrication}
With our machine setup, we fabricate in total $\Pi = 10$ parts divided into three complex and seven simple geometries. All specimens are manufactured under argon atmosphere in an \textit{EOS M 290} machine (\textit{EOS GmbH, Germany}) with a \SI{400}{\watt} fiber laser. For this study, a gas-atomized (\textit{argon}) \textit{AlSi10Mg} powder (\textit{Eckart TLS GmbH, Germany}) was applied. Each specimen was built with rotating the laser scanning direction \ang{67} after each layer. 
For this study, we used different processing parameters for both types of geometries. Our complex geometry was fabricated with a parameter configuration of laser power of $P = \SI{370}{\watt}$, scan speed $v = \SI{1300}{\milli\meter\per\second}$, hatch distance $h = \SI{180}{\micro\meter}$ at part positions $p = 1, 2, 3$. This was done to evaluate the influence of geometry features on model prediction performance. With a constant layer thickness $t = \SI{30}{\micro\meter}$, the energy density of $E_v = \SI{49.93e9}{\joule\per\meter\cubed}$ is calculated using the following equation (\ref{eq:energy_density}), which quantifies the amount of energy generated per volume of material during the scanning of a layer.

\begin{equation}
\label{eq:energy_density}
    \mathit{E_v} = \frac{P}{v h t} 
\end{equation}

To evaluate the influence of parameter configuration on model prediction performance, each specimen of the simple geometry, at part position $p = 4, \dots, 10$, was fabricated with a different parameter configuration. For this purpose, the laser power has been adjusted to a range of $340$ to $\SI{390}{\watt}$, the scan speed has been set between $700$ and $\SI{1300}{\milli\meter\per\second}$, and the hatch distance has been varied between 160 and $\SI{210}{\micro\meter}$. This has resulted in an energy density range between \num{41.51e9} and $\SI{123.80e9}{\joule\per\meter\cubed}$. We summarized all parameter configurations in \autoref{tab:Parameter_configuration}. As each parameter configuration allows for a distinct part position to be allocated, $p$ is equally used for both interchangeably.  

\begin{table}[htb]
\centering
\caption{PBF-LB/M process parameters for all parts $p$ used}
\label{tab:Parameter_configuration}
\begin{tabular}{c *{5}{c}}
\toprule
$p$ & 
$P [\si{\watt}]$ & 
$v [\si{\milli\meter\per\second}]$ & 
$h [\si{\micro\meter}]$ & 
$t [\si{\micro\meter}]$ &
$E_v [\si{\joule\per\meter\cubed}]$ \\ 
\midrule
1-3 & 370 & 1300   & 190 & 30   & \num{49.93e9} \\
4   & 340 & 1300   & 210 & 30   & \num{41.51e9}  \\
5   & 370 & 1300   & 190 & 30   & \num{49.93e9}  \\
6   & 340 & 1000   & 190 & 30   & \num{59.64e9}  \\
7   & 370 & 900    & 210 & 30   & \num{65.25e9}  \\
8   & 370 & 700    & 190 & 30   & \num{72.12e9}  \\
9   & 370 & 700    & 210 & 30   & \num{83.90e9}  \\
10  & 390 & 700    & 160 & 30   & \num{123.80e9}  \\
\bottomrule
\end{tabular}
\end{table}

\subsection{XCT measurements}
\label{subsec:XCT_measurements}
The samples were removed from the build plate with a rotating tool at a distance of approximately \SI{0.5}{\milli\meter}, which indicates the thickness of the tool blade. All parts scanned using an X-ray computer tomograph (CT) \textit{Werth Tomoscope HV Compact}. The system features a micro-focus transmission tube with a capacity of up to \SI{225}{\kilo\volt}. The detector size covers $410 \times \SI{410}{\milli\meter}$ with a resolution of $2048 \times \SI{2048}{px}$. The scanning result is a volumetric representation of the specimen in a voxel-based format, where the measured voxel density corresponds to grayscale values. We automatically detect pore locations within the part using the integrated threshold method from the commercial software \textit{VGStudio}. With the configuration of intensity threshold $t = 11500$ and minimum pore diameter of \SI{10}{\milli\meter} the software calculates pore locations, and highlights them inside the scanned volume.

\subsection{Data processing and labelling}
\label{subsec:data_labeling}
With the monitoring system setup and the acquired raw data, we apply preprocessing steps to create a dataset for model training.   
We first manually crop out all parts at their positions within both image modalities (HR and OT). We determine the rectangular cross-section boundary of each part, by selecting four reference points per part. We crop each image along the rectangular bounding box and store them as separated image files with machine parameter configuration $p$ and layer $l$. Although we fabricated two different geometries, each geometry can be allocated to a parameter configuration unambiguously, thus we dismiss a distinction in the mathematical notation. This results for each image modality $m$ in a sequence of images $\mathcal{X} = \{x^m_l \mid l = 1 \dots \Lambda\}$ with the total number of layers $\Lambda = 712$.

To match the CT image data with the OT and HR data, we first rotate the 3D CT volume so that it matches the orientation of the printing direction. Next, we slice the volume from the position of the last layer (top) to that of the first layer (bottom) with the corresponding layer thickness of $t=\SI{30}{\micro\meter}$. Afterward, we similarly crop all CT cross-section images at the same reference points as the OT and HR images. After cropping the CT images, we determine, based on a list of pixel positions that were identified by \textit{VGStudio} as pores, the probability distribution using equation (\ref{eq:kde}). 

\begin{figure}[htb]
    \centering
    \includegraphics[width=\textwidth]{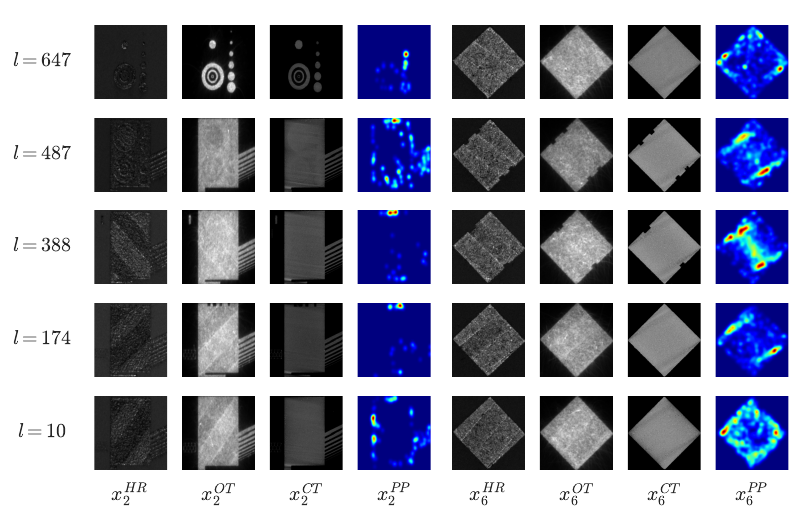}
    \caption{Samples of the training dataset, showing images of two parts $p=2$ and $p=6$. Five exemplary layers are shown for each part, with their corresponding data modality HR, OT, CT, and PP arranged from left to right respectively. For the pore probability distribution, blue indicates low and red high probability values.}
    \label{fig:dataset}
\end{figure}

Our data processing and labelling steps result in a dataset $\mathcal{D}$ depicted in the following equation (\ref{eq:dataset}) with a total number of $T=7112$ triplets containing HR images $\mathcal{X}^{HR}$, OT images $\mathcal{X}^{OT}$, and corresponding pore probability images $\mathcal{X}^{PP}$ for each parameter configuration. For illustration purposes, a sample from $\mathcal{D}$ is depicted in \autoref{fig:dataset}. 

\begin{equation}
\label{eq:dataset}
    \mathcal{D} = \left\{\left(\mathcal{X}^{HR}, \mathcal{X}^{OT}, \mathcal{X}^{PP}\right)_{p, l} \mid 1 \leq p \leq \Pi, 1 \leq l \leq \Lambda \right\}
\end{equation}

\subsection{Pore Probability Prediction with Segmentation Models}
\label{subsec:ppp_with_seg}
With the given dataset containing OT, HR, and Pore Probability images for each layer and part, we aim to train a model in a supervised fashion that describes the functional relationship between monitoring images and pore probability occurrence. 
Due to their comparability in task difficulty, we frame the pore location estimation task as an image segmentation problem — that is, a pixel-wise classification of an image —, where the segmentation masks are derived from Pore Probability images. 
However, instead of classifying each pixel into a discrete class, we train different state-of-the-art segmentation models to estimate the pore occurrence probability at each pixel. 
This task can be expressed using the objective function
\begin{equation}
    \label{eq:ppp}
    \underset{\theta}{\arg\min} \, \mathit{L} \left( f_{\theta} \left(x^{HR}, x^{OT}\right), x^{PP} \right),
\end{equation}
 where we search weights $\theta$ of a segmentation model $f$, that minimizes the loss function 
 \begin{equation}
    \label{eq:loss}
    \displaystyle L = \sum^{\mathcal{D}_t}|x-y|
\end{equation}
known as mean absolute error (MAE) for a separated test set $\mathcal{D}_t \subset \mathcal{D}$ of size $20 \%$. With our objective defined, we execute a hyperparameter tuning strategy, using Bayesian optimization to find the best hyperparameter configuration within the ranges of learning rate $\alpha = [10^{-6}, 10^{-3}]$ and batch size $\beta = \{16, 32, 64\}$. For each model architecture, we run $50$ trials with $60$ epochs and report the MAE calculated on a separate validation set $\mathcal{D}_v \subset \mathcal{D}$ of size $20 \%$. Afterward, we train ten models with different random seeds for each architecture, with the best hyperparameter configuration.

Semantic segmentation is an important computer vision problem for a variety of applications. It can be formulated as a classification problem on pixel-level, where each pixel is labelled into a set of categories. Deep Learning models in particular have shown promising results in solving the segmentation task \cite{Minaee.2022}. Consequently, deep learning models were also applied for segmenting pores within different monitoring modalities to accurately localize pore positions (cf. section \ref{sec:introduction}). Because previous work has not shown success in directly segmenting pores within in-situ monitoring data using segmentation models, we demonstrate the effectiveness of our approach by training different segmentation models to estimate the probability of pore occurrence.
Within our experiments, we test the following popular semantic segmentation models provided by the library \textit{Segmentation Models} \cite{Iakubovskii.2019}:
\begin{enumerate}
    \item Unet \cite{Ronneberger.2015}. The network uses multiple convolutional operations for downsampling and upsampling. The upsampling operations input is the concatenation of downsampling outputs and upsampling outputs.
    \item FPN \cite{Lin.2017}. The main component of the architecture is a feature pyramid extracted from hierarchical convolutional neural networks.
    \item LinkNet \cite{Chaurasia.2017}. The architecture is optimized regarding several weights. By combining different convolutional operations during downsampling and upsampling.
    \item DeepLabV3 \cite{Chen.2017}. Uses dilated convolution in parallel to incorporate global context information into the prediction.
    \item MAnet \cite{Fan.2020}. Initially applied in the medical field, the Multi-scale Attention Net uses the self-attention mechanism to integrate local features with their global dependencies into the segmentation prediction. 
\end{enumerate}


\section{Results}
\label{sec:results}
\begin{figure}[htb]
    \centering
    \includegraphics[width=\textwidth]{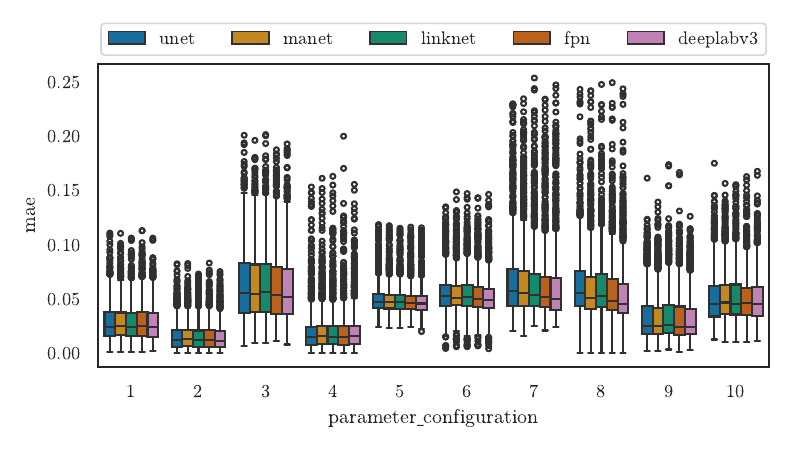}
    \caption{Box plots of MAE for all experiments. Model architectures are color-coded. The circles above the whiskers can be considered outliers. The smaller the MAE the better the model performance.}
    \label{fig:MAE_experimenets}
\end{figure}

Based on our geometry, we aimed to investigate the prediction performance of different model architectures dependent on (i) geometry features and (ii) process parameters. First, we give a general overview of different model architecture performances in  \autoref{fig:MAE_experimenets}. All five models showcase comparable performance over all experiments. Only deeplabv3 performs slightly better at the experiments $p = 3, 6, 7, 8$. Otherwise, no model significantly outperforms the other ones, in which the prediction error median of one model is outside the lower and upper quartile of the other models. However, the model prediction error from $p = 3$, that was fabricated with the same parameters and geometry as $p = 1, 2$, showcases significant differences in model performance. Thus, indicating a potential systematic error during data preprocessing or model training. 

Between the XYZ cube experiments, the prediction error deviation is smaller for part positions $p = 4, 5, 6$ compared to the other experiments, indicating a more robust model performance. Furthermore, for all models, many outliers are observable by the dots outside the whiskers of the box plot. This demonstrates that although the models show on average a sufficient prediction performance, for a significant number of instances the performance drops significantly.  

\begin{figure}[htb]
    \centering
    \includegraphics[width=\textwidth]{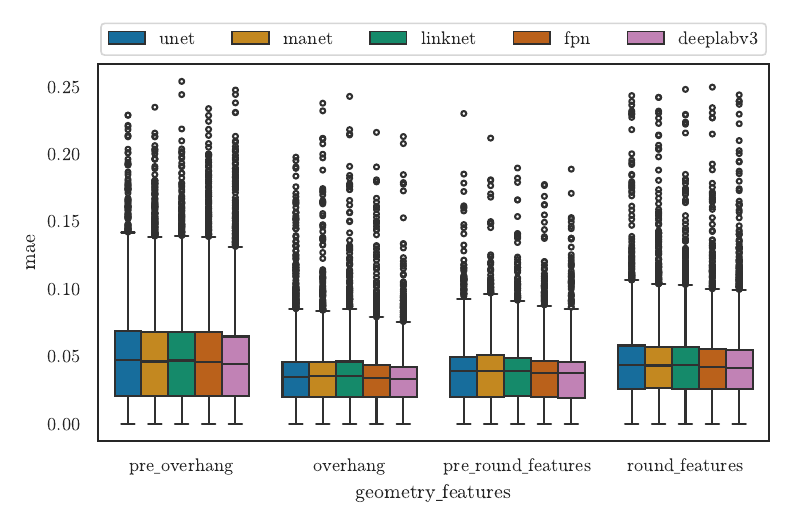}
    \caption{Box plots of MAE values for each model over  Geometry Features derived from the complex geometry. Geometry features are derived from layers with different geometry features.}
    \label{fig:geometry_features}
\end{figure}

To get a better understanding, we also investigated the influence of geometry features ($p = 1, 2 ,3$) on prediction performance. First, we split up the complex geometries layer-wise based on their visible features into four sections \textit{pre overhang}, \textit{overhang}, \textit{pre round features}, and \textit{round features}. The first section showcases the area below the overhang features, where only angular slits are visible with layers $l = 1 \dots 245$. The second section contains all layers $l = 246 \dots 430$ with overhang features. The third section includes all layers $l = 431 \dots 487$ with neither overhang nor slit features, and the final section contains all layers $l = 488 \dots 712$ with round features, like varying cylinders and the semi-sphere. The MAE over these four sections is displayed in \autoref{fig:geometry_features}. From the figure, we derive again that no model significantly outperforms the other models over different geometry features. Furthermore, the performance between each geometry feature class does not alter much, however, the performance for overhang and pre overhang shows the most stable prediction performance, based on lower variance and slightly lower median MAE.

\begin{figure}[p]
    \centering
    \includegraphics[width=\textwidth]{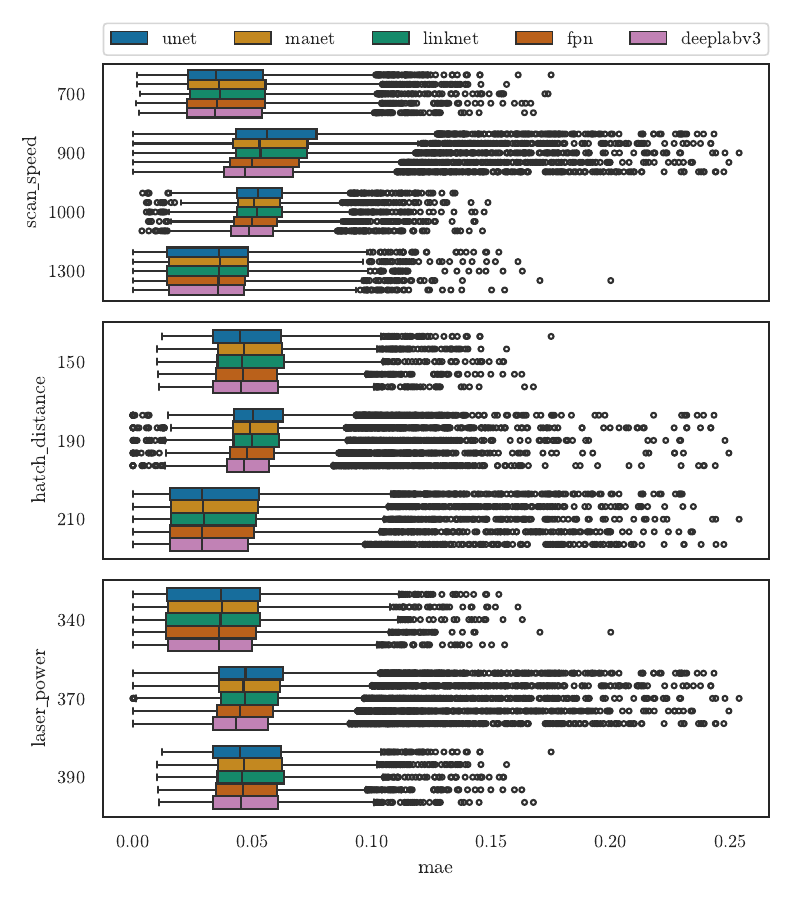}
    \caption{Box plot of MAE depicted dependent on different process parameters. The results are derived solely from the results of the XYZ cube. Model architectures are color-coded.}
    \label{fig:parameter_influence}
\end{figure}

Besides geometry features, we investigated the influence of machine parameters on performance prediction using the resulting MAE from the XYZ cube ($p = 4, \dots, 10$). Because each cube was printed with different but also overlapping process parameters (cf. \autoref{tab:Parameter_configuration}), we looked at each parameter individually and compared their model performance based on the MAE in \autoref{fig:parameter_influence}. Compared to the geometry features, process parameters have significantly higher influence on mode performance. Increasing hatch distance, and reducing laser power results in lower MAE values. With increasing scan speed, the model performance first decreases and increases again. Deeplabv3 overall showcases the lowest prediction errors over all parameter configurations. Changing parameter configuration has a bigger impact on the resulting image contrast compared to changes in geometry. Therefore, it is plausible that the changing parameter configuration has a higher impact on model performance.

\begin{figure}[p]
    \centering
    \includegraphics[width=\textwidth]{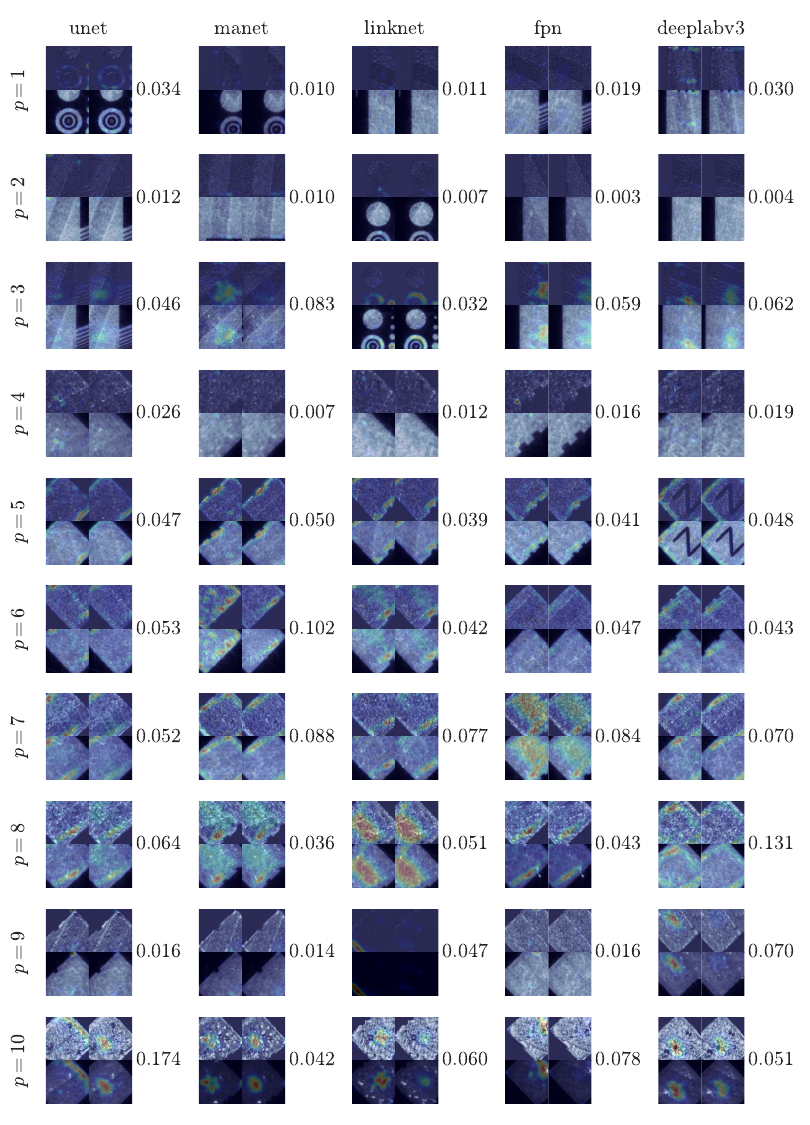}
    \caption{Prediction performance of each experiment and model is displayed with corresponding MAE values.     
    Within an image grid, the data types HR (top) and OT (bottom) are overlapped with predetermined (left) and estimated (right) pore probability.}
    \label{fig:segmentation_visuals}
\end{figure}

Finally, we present visually the predicted probability distribution over all experiments and models in \autoref{fig:segmentation_visuals}. Low MAE values below around $0.05$ indicate qualitatively accurate pore prediction, where the predicted regions overlap highly with the estimated probability distribution. Observable failure modes occur if the model fails to predict any probability distribution, or if the probability shape or location is highly misaligned. As can be seen in the figure, most pore predictions showcase a high overlap compared to the predetermined probability distribution. However, as shown in the figure for $\left(p = 3, \text{manet} \right), \left(p = 9, \text{deeplabv3} \right), \left(p = 10, \text{fpn} \right)$ nearly no pore occurrence were predicted, showcasing the first failure mode. The samples $\left(p=3, \text{deeplabv3}\right), \left(p=10, \text{unet}\right)$ showcase the second failure mode, where the predicted probability distribution is highly misaligned compared to the predefined probability distribution. With the remaining probability distributions showing a good overlap between prediction and predetermined probability distributions, our results demonstrate the performance of localizing pores with segmentation models.

\section{Discussion}
\label{sec:discussion}
The reformulation of the pore segmentation problem to a pore estimation problem using KDE, allowed us to train segmentation models to learn the pore probability occurrence from in-situ monitoring data. Our results indicate, that all segmentation models show comparable performance in accurately estimating pore occurrence. 
However, within our work, we did not verify the true pore position at the location determined by the CT software solution. For our study, we assumed that the selected threshold would result in ground truth pore positions, which is an overestimation of pore occurrence. Thus leading to a higher likelihood prediction of pores than the actual occurrence of pores. Currently, the most reliable source of investigating pores is by opening the specimen and analyzing the resulting cross-section. As long as measuring defect occurrence does not become more accurate, the model performance also stays limited in that regard. We are certain, that for future work improving the labelling process of determining pore locations (or other defects) in post-processing will automatically lead to more reliable model predictions.

Additionally, another error-source lies in the layer-wise mapping of different data modalities, especially during the mapping process between in-situ monitoring data and CT cross-section images. Here, two potential misalignment errors are possible, first the CT volume rotation may not match with the captured in-situ image. An automatic solution that matches the orientation using markers may reduce potential human errors. Second, and maybe more relevant, is an accurate alignment of layer positions and thus the occurrence of pores. Although we used geometry features to determine corresponding layer pairs, a perfect alignment with the required accuracy, which guarantees pore occurrence in the observed monitoring data, is not given within our study. Therefore, further studies investigating potential labelling errors are needed; A representative benchmark dataset where the data quality was approved could benefit Machine Learning practitioners in AM.

Besides the aspect of data quality, other modelling approaches of the probability distribution are worth investigating. We chose KDE with a Gaussian kernel for demonstration and simplicity purposes, but other probability distribution estimators may be more suitable for estimating pore location. Certainly, the selection of the kernel and bandwidth could be investigated for improvement, and other concepts from generative models could be applied to further learn a more accurate pore probability distribution.

In our study, we evaluated model performance based on geometry features and machine parameters. Further investigations, that either stress model performance like real-world fabrication jobs with high defect probability or controlled studies where the appearance of pores is more deterministic, could give a better understanding of model performance and improvements. We believe that with ever-increasing sensor technology and an extension from single layer predictions to future layer predictions will enable new control strategies preventing defects from forming.

\section{Conclusion}
\label{sec:conclusion}
This work demonstrated how to enable pore localization within a layer of in-situ monitoring data of \lpbf processes; That is, by estimating the pore location using KDE, segmentation models could be used to predict the probability of pore occurrence. From our work, the following conclusions are drawn:
\begin{itemize}
    \item The reformulation of pore segmentation into a pore estimation problem using KDE is demonstrated to provide an estimation of pore occurrence. The results show that the segmentation models used provide comparable performance in estimating pore occurrence.
    \item Our experiments have shown that changing machine parameters have a greater influence on model performance compared to geometry features. Thus, our approach performs better, when the machine parameters are kept within a certain range. 
    \item Within our experiments, no segmentation model could significantly outperform the other ones, indicating potential model performance improvements, by tailoring the models to the use case of pore probability prediction.
    \item Our work has showcased how real-time in-situ monitoring data can be utilized to predict the occurrence of pores. We demonstrated an accessible methodology for training segmentation models to translate non-CT data into an estimation of pore probability.
\end{itemize}

\section*{Statements and Declarations}
\label{sec:Declaration of Competing Interest }
The authors declare that they have no known competing financial
interests or personal relationships that could have appeared to influence
the work reported in this paper. 

\section*{Data Availability}
\label{sec:Data Availability}
Data will be made available on request.



\bibliography{Manuscript}

\begin{thebibliography}{10}
\providecommand{\doi}[1]{\url{https://doi.org/#1}}
\bibcommenthead

\bibitem[\protect\citeauthoryear{Arif et~al.}{2023}]{Arif.2023}
Arif ZU, Khalid MY, Noroozi R, Hossain M, Shi HH, Tariq A, et~al.
\newblock Additive manufacturing of sustainable biomaterials for biomedical applications.
\newblock Asian journal of pharmaceutical sciences. 2023;18(3):100812.
\newblock \doi{10.1016/j.ajps.2023.100812}.

\bibitem[\protect\citeauthoryear{Blakey-Milner et~al.}{2021}]{BlakeyMilner.2021}
Blakey-Milner B, Gradl P, Snedden G, Brooks M, Pitot J, Lopez E, et~al.
\newblock Metal additive manufacturing in aerospace: A review.
\newblock Materials {\&} Design. 2021;209:110008.
\newblock \doi{10.1016/j.matdes.2021.110008}.

\bibitem[\protect\citeauthoryear{Grasso et~al.}{2021}]{Grasso.2021}
Grasso M, Remani A, Dickins A, Colosimo BM, Leach RK.
\newblock In-situ measurement and monitoring methods for metal powder bed fusion: an updated review.
\newblock Measurement Science and Technology. 2021;32(11):112001.
\newblock \doi{10.1088/1361-6501/ac0b6b}.

\bibitem[\protect\citeauthoryear{Behery et~al.}{2023}]{Behery.2023}
Behery M, Brauner P, Zhou HA, Uysal MS, Samsonov V, Bellgardt M, et~al.
\newblock Actionable Artificial Intelligence for the Future of Production.
\newblock In: Padberg M, Piller FT, Jarke M, {van der Aalst} W, Schuh G, Brecher C, editors. Internet of Production. Interdisciplinary Excellence Accelerator Series. Cham: {Springer International Publishing}; 2023. p. 1--46.

\bibitem[\protect\citeauthoryear{Liebenberg and Jarke}{2023}]{Liebenberg.2023}
Liebenberg M, Jarke M.
\newblock Information systems engineering with Digital Shadows: Concept and use cases in the Internet of Production.
\newblock Information Systems. 2023;114:102182.
\newblock \doi{10.1016/j.is.2023.102182}.

\bibitem[\protect\citeauthoryear{Sanaei and Fatemi}{2021}]{Sanaei.2021}
Sanaei N, Fatemi A.
\newblock Defects in additive manufactured metals and their effect on fatigue performance: A state-of-the-art review.
\newblock Progress in Materials Science. 2021;117:100724.
\newblock \doi{10.1016/j.pmatsci.2020.100724}.

\bibitem[\protect\citeauthoryear{Shevchik et~al.}{2018}]{Shevchik.2018}
Shevchik SA, Kenel C, Leinenbach C, Wasmer K.
\newblock Acoustic emission for in situ quality monitoring in additive manufacturing using spectral convolutional neural networks.
\newblock Additive Manufacturing. 2018;21:598--604.
\newblock \doi{10.1016/j.addma.2017.11.012}.

\bibitem[\protect\citeauthoryear{Khanzadeh et~al.}{2018}]{Khanzadeh.2018}
Khanzadeh M, Chowdhury S, Marufuzzaman M, Tschopp MA, Bian L.
\newblock Porosity prediction: Supervised-learning of thermal history for direct laser deposition.
\newblock Journal of Manufacturing Systems. 2018;47:69--82.
\newblock \doi{10.1016/j.jmsy.2018.04.001}.

\bibitem[\protect\citeauthoryear{Gobert et~al.}{2018}]{Gobert.2018}
Gobert C, Reutzel EW, Petrich J, Nassar AR, Phoha S.
\newblock Application of supervised machine learning for defect detection during metallic powder bed fusion additive manufacturing using high resolution imaging.
\newblock Additive Manufacturing. 2018;21:517--528.
\newblock \doi{10.1016/j.addma.2018.04.005}.

\bibitem[\protect\citeauthoryear{Zhang et~al.}{2019}]{Zhang.2019}
Zhang B, Liu S, Shin YC.
\newblock In-Process monitoring of porosity during laser additive manufacturing process.
\newblock Additive Manufacturing. 2019;28:497--505.
\newblock \doi{10.1016/j.addma.2019.05.030}.

\bibitem[\protect\citeauthoryear{Seifi et~al.}{2019}]{Seifi.2019}
Seifi SH, Tian W, Doude H, Tschopp MA, Bian L.
\newblock Layer-Wise Modeling and Anomaly Detection for Laser-Based Additive Manufacturing.
\newblock Journal of Manufacturing Science and Engineering. 2019;141(8).
\newblock \doi{10.1115/1.4043898}.

\bibitem[\protect\citeauthoryear{Imani et~al.}{2019}]{Imani.2019}
Imani F, Chen R, Diewald E, Reutzel E, Yang H.
\newblock Deep Learning of Variant Geometry in Layerwise Imaging Profiles for Additive Manufacturing Quality Control.
\newblock Journal of Manufacturing Science and Engineering. 2019;141(11).
\newblock \doi{10.1115/1.4044420}.

\bibitem[\protect\citeauthoryear{Gaikwad et~al.}{2019}]{Gaikwad.2019}
Gaikwad A, Imani F, Yang H, Reutzel E, Rao P.
\newblock In Situ Monitoring of Thin-Wall Build Quality in Laser Powder Bed Fusion Using Deep Learning.
\newblock Smart and Sustainable Manufacturing Systems. 2019;3(1):20190027.
\newblock \doi{10.1520/SSMS20190027}.

\bibitem[\protect\citeauthoryear{Caggiano et~al.}{2019}]{Caggiano.2019}
Caggiano A, Zhang J, Alfieri V, Caiazzo F, Gao R, Teti R.
\newblock Machine learning-based image processing for on-line defect recognition in additive manufacturing.
\newblock CIRP Annals. 2019;68(1):451--454.
\newblock \doi{10.1016/j.cirp.2019.03.021}.

\bibitem[\protect\citeauthoryear{Kwon et~al.}{2020}]{Kwon.2020}
Kwon O, Kim HG, Ham MJ, Kim W, Kim GH, Cho JH, et~al.
\newblock A deep neural network for classification of melt-pool images in metal additive manufacturing.
\newblock Journal of Intelligent Manufacturing. 2020;31(2):375--386.
\newblock \doi{10.1007/s10845-018-1451-6}.

\bibitem[\protect\citeauthoryear{Gobert et~al.}{2020}]{Gobert.2020}
Gobert C, Kudzal A, Sietins J, Mock C, Sun J, McWilliams B.
\newblock Porosity segmentation in X-ray computed tomography scans of metal additively manufactured specimens with machine learning.
\newblock Additive Manufacturing. 2020;36:101460.
\newblock \doi{10.1016/j.addma.2020.101460}.

\bibitem[\protect\citeauthoryear{Snow et~al.}{2021}]{Snow.2021}
Snow Z, Diehl B, Reutzel EW, Nassar A.
\newblock Toward in-situ flaw detection in laser powder bed fusion additive manufacturing through layerwise imagery and machine learning.
\newblock Journal of Manufacturing Systems. 2021;59:12--26.
\newblock \doi{10.1016/j.jmsy.2021.01.008}.

\bibitem[\protect\citeauthoryear{Petrich et~al.}{2021}]{Petrich.2021}
Petrich J, Snow Z, Corbin D, Reutzel EW.
\newblock Multi-modal sensor fusion with machine learning for data-driven process monitoring for additive manufacturing.
\newblock Additive Manufacturing. 2021;48:102364.
\newblock \doi{10.1016/j.addma.2021.102364}.

\bibitem[\protect\citeauthoryear{Zhang et~al.}{2022}]{Zhang.2022}
Zhang J, Lyu T, Hua Y, Shen Z, Sun Q, Rong Y, et~al.
\newblock Image Segmentation for Defect Analysis in Laser Powder Bed Fusion: Deep Data Mining of X-Ray Photography from Recent Literature.
\newblock Integrating Materials and Manufacturing Innovation. 2022;11(3):418--432.
\newblock \doi{10.1007/s40192-022-00272-5}.

\bibitem[\protect\citeauthoryear{Yang et~al.}{2022}]{Yang.2022}
Yang H, Wang W, Li C, Qi J, Wang P, Lei H, et~al.
\newblock Deep learning-based X-ray computed tomography image reconstruction and prediction of compression behavior of 3D printed lattice structures.
\newblock Additive Manufacturing. 2022;54:102774.
\newblock \doi{10.1016/j.addma.2022.102774}.

\bibitem[\protect\citeauthoryear{Snow et~al.}{2022}]{Snow.2022}
Snow Z, Reutzel EW, Petrich J.
\newblock Correlating in-situ sensor data to defect locations and part quality for additively manufactured parts using machine learning.
\newblock Journal of Materials Processing Technology. 2022;302:117476.
\newblock \doi{10.1016/j.jmatprotec.2021.117476}.

\bibitem[\protect\citeauthoryear{Pandiyan et~al.}{2022}]{Pandiyan.2022}
Pandiyan V, Masinelli G, Claire N, Le-Quang T, Hamidi-Nasab M, de~Formanoir C, et~al.
\newblock Deep learning-based monitoring of laser powder bed fusion process on variable time-scales using heterogeneous sensing and operando X-ray radiography guidance.
\newblock Additive Manufacturing. 2022;58:103007.
\newblock \doi{10.1016/j.addma.2022.103007}.

\bibitem[\protect\citeauthoryear{Nemati et~al.}{2022}]{Nemati.2022}
Nemati S, Ghadimi H, Li X, Butler LG, Wen H, Guo S.
\newblock Automated Defect Analysis of Additively Fabricated Metallic Parts Using Deep Convolutional Neural Networks.
\newblock Journal of Manufacturing and Materials Processing. 2022;6(6):141.
\newblock \doi{10.3390/jmmp6060141}.

\bibitem[\protect\citeauthoryear{Ansari et~al.}{2022}]{Ansari.2022}
Ansari MA, Crampton A, Garrard R, Cai B, Attallah M.
\newblock A Convolutional Neural Network (CNN) classification to identify the presence of pores in powder bed fusion images.
\newblock The International Journal of Advanced Manufacturing Technology. 2022;120(7-8):5133--5150.
\newblock \doi{10.1007/s00170-022-08995-7}.

\bibitem[\protect\citeauthoryear{Ye et~al.}{2023}]{Ye.2023}
Ye J, Poudel A, Liu J, Vinel A, Silva D, Shao S, et~al.
\newblock Machine learning augmented X-ray computed tomography features for volumetric defect classification in laser beam powder bed fusion.
\newblock The International Journal of Advanced Manufacturing Technology. 2023;126(7-8):3093--3107.
\newblock \doi{10.1007/s00170-023-11281-9}.

\bibitem[\protect\citeauthoryear{Surana et~al.}{2023}]{Surana.2023}
Surana A, Lynch ME, Nassar AR, Ojard GC, Fisher BA, Corbin D, et~al.
\newblock Flaw Detection in Multi-Laser Powder Bed Fusion Using In Situ Coaxial Multi-Spectral Sensing and Deep Learning.
\newblock Journal of Manufacturing Science and Engineering. 2023;145(5).
\newblock \doi{10.1115/1.4056540}.

\bibitem[\protect\citeauthoryear{Gorgannejad et~al.}{2023}]{Gorgannejad.2023}
Gorgannejad S, Martin AA, Nicolino JW, Strantza M, Guss GM, Khairallah S, et~al.
\newblock Localized keyhole pore prediction during laser powder bed fusion via multimodal process monitoring and X-ray radiography.
\newblock Additive Manufacturing. 2023;78:103810.
\newblock \doi{10.1016/j.addma.2023.103810}.

\bibitem[\protect\citeauthoryear{Pan et~al.}{2024}]{Pan.2024}
Pan J, Hu D, Zhou L, {Di Huang}, Wang Y, Wang R.
\newblock Semantic segmentation of defects based on DCNN and its application on fatigue lifetime prediction for SLM Ti-6Al-4V alloy.
\newblock Philosophical Transactions of the Royal Society A: Mathematical, Physical and Engineering Sciences. 2024;382(2264):20220396.
\newblock \doi{10.1098/rsta.2022.0396}.

\bibitem[\protect\citeauthoryear{Dong et~al.}{2024}]{Dong.2024}
Dong W, Lian J, Yan C, Zhong Y, Karnati S, Guo Q, et~al.
\newblock Deep-Learning-Based Segmentation of Keyhole in In-Situ X-ray Imaging of Laser Powder Bed Fusion.
\newblock Materials (Basel, Switzerland). 2024;17(2).
\newblock \doi{10.3390/ma17020510}.

\bibitem[\protect\citeauthoryear{Desrosiers et~al.}{2024}]{Desrosiers.2024}
Desrosiers C, Letenneur M, Bernier F, Pich{\'e} N, Provencher B, Cheriet F, et~al.
\newblock Automated porosity segmentation in laser powder bed fusion part using computed tomography: a validity study.
\newblock Journal of Intelligent Manufacturing. 2024;p. 1--21.
\newblock \doi{10.1007/s10845-023-02296-w}.

\bibitem[\protect\citeauthoryear{Fuchs and Eischer}{2020}]{Fuchs.2020}
Fuchs L, Eischer C.: In-process monitoring systems for metal additive manufacturing.
\newblock Available from: \url{https://www.eos-apac.info/upload/2020-07/159522956575650000.pdf}.

\bibitem[\protect\citeauthoryear{Zenzinger et~al.}{2015}]{Zenzinger.2015}
Zenzinger G, Bamberg J, Ladewig A, Hess T, Henkel B, Satzger W.
\newblock Process monitoring of additive manufacturing by using optical tomography.
\newblock In: AIP Conference Proceedings; 2015. p. 164--170.

\bibitem[\protect\citeauthoryear{Yoder et~al.}{2019}]{Yoder.2019}
Yoder S, Nandwana P, Paquit V, Kirka M, Scopel A, Dehoff RR, et~al.
\newblock Approach to qualification using E-PBF in-situ process monitoring in Ti-6Al-4V.
\newblock Additive Manufacturing. 2019;28:98--106.
\newblock \doi{10.1016/j.addma.2019.03.021}.

\bibitem[\protect\citeauthoryear{G{\"o}gelein et~al.}{2018}]{Gogelein.2018}
G{\"o}gelein A, Bamberg J, Zenzinger G, Ladewig A.: Process Monitoring of Additive Manufacturing by Using Optical Tomography.
\newblock Unpublished.

\bibitem[\protect\citeauthoryear{Minaee et~al.}{2022}]{Minaee.2022}
Minaee S, Boykov Y, Porikli F, Plaza A, Kehtarnavaz N, Terzopoulos D.
\newblock Image Segmentation Using Deep Learning: A Survey.
\newblock IEEE transactions on pattern analysis and machine intelligence. 2022;44(7):3523--3542.
\newblock \doi{10.1109/TPAMI.2021.3059968}.

\bibitem[\protect\citeauthoryear{Iakubovskii}{2019}]{Iakubovskii.2019}
Iakubovskii P.: Segmentation Models Pytorch.
\newblock GitHub.

\bibitem[\protect\citeauthoryear{Ronneberger et~al.}{2015}]{Ronneberger.2015}
Ronneberger O, Fischer P, Brox T.: U-Net: Convolutional Networks for Biomedical Image Segmentation.
\newblock Available from: \url{http://arxiv.org/pdf/1505.04597.pdf}.

\bibitem[\protect\citeauthoryear{Lin et~al.}{2017}]{Lin.2017}
Lin TY, Dollar P, Girshick R, He K, Hariharan B, Belongie S.
\newblock Feature Pyramid Networks for Object Detection.
\newblock In: 30th IEEE Conference on Computer Vision and Pattern Recognition. Piscataway, NJ: IEEE; 2017. p. 936--944.

\bibitem[\protect\citeauthoryear{Chaurasia and Culurciello}{2017}]{Chaurasia.2017}
Chaurasia A, Culurciello E.: LinkNet: Exploiting encoder representations for efficient semantic segmentation.
\newblock Available from: \url{http://arxiv.org/pdf/1707.03718}.

\bibitem[\protect\citeauthoryear{Chen et~al.}{2017}]{Chen.2017}
Chen LC, Papandreou G, Schroff F, Adam H.: Rethinking Atrous Convolution for Semantic Image Segmentation.
\newblock Available from: \url{http://arxiv.org/pdf/1706.05587}.

\bibitem[\protect\citeauthoryear{Fan et~al.}{2020}]{Fan.2020}
Fan T, Wang G, Li Y, Wang H.
\newblock MA-Net: A Multi-Scale Attention Network for Liver and Tumor Segmentation.
\newblock IEEE Access. 2020;8:179656--179665.
\newblock \doi{10.1109/ACCESS.2020.3025372}.

\end{thebibliography}

\end{document}